\documentclass{article}

\usepackage[preprint, nonatbib]{template}

\usepackage[utf8]{inputenc} 
\usepackage[T1]{fontenc}    
\usepackage{hyperref}       
\usepackage{url}            
\usepackage{booktabs}       
\usepackage{amsfonts}       
\usepackage{nicefrac}       
\usepackage{microtype}      
\usepackage{xcolor}         
\usepackage{graphicx}
\usepackage{amsmath}

\title{Fine-tuning of Large Language Models for Domain-Specific Cybersecurity Knowledge}

\author{%
  Yuan Huang \\
  Shenzhen College of International Education \\
  Shenzhen, China 518043 \\
  \texttt{s22404.huang@stu.scie.com.cn} \\
}

\begin{document}

\maketitle

\begin{abstract}
Recent advancements in training paradigms for Large Language Models (LLMs) have unlocked their remarkable capabilities in natural language processing and cross-domain generalization. While LLMs excel in tasks like programming and mathematical problem-solving, their zero-shot performance in specialized domains requiring expert knowledge, such as cybersecurity, is often suboptimal. This limitation arises because foundational LLMs are designed for general-purpose applications, constraining their ability to encapsulate domain-specific expertise within their parameter space. To address this, we explore fine-tuning strategies to embed cybersecurity knowledge into LLMs, enhancing their performance in cybersecurity question-answering (Q\&A) tasks while prioritizing computational efficiency. Specifically, we investigate Supervised Fine-Tuning (SFT), Low-Rank Adaptation (LoRA), and Quantized Low-Rank Adaptation (QLoRA) using a cybersecurity Q\&A dataset. Our results demonstrate that these fine-tuning approaches significantly outperform the foundational model in cybersecurity Q\&A tasks. Moreover, LoRA and QLoRA achieve comparable performance to SFT with substantially lower computational costs, offering an efficient pathway for adapting LLMs to specialized domains. Our work highlights the potential of low-rank fine-tuning strategies to bridge the gap between general-purpose LLMs and domain-specific applications.
\end{abstract}

\section{Introduction}

The transformer architecture \cite{vaswani2017attention}, scaling laws \cite{kaplan2020scaling}, and advanced training pipelines involving large-scale pre-training, supervised post-training, and alignment \cite{achiam2023gpt} have significantly enhanced the performance of Large Language Models (LLMs). These advancements enable LLMs to excel in diverse downstream tasks, such as conversational question-answering (Q\&A)\cite{abbasiantaeb2024let}, knowledge graph construction \cite{carta2023iterative}, and reasoning \cite{kojima2022large}, often in a zero-shot setting.

However, LLMs often underperform in specialized domains requiring expert knowledge when used without adaptation. This limitation stems from the information compression inherent in the transformer architecture \cite{huang2024compression}. Due to hardware constraints on training and inference, foundational LLMs have a finite number of parameters, restricting the amount of knowledge that can be encoded. As general-purpose models, off-the-shelf LLMs primarily capture broad, general knowledge, which limits their effectiveness in niche domains like cybersecurity.

To address this, fine-tuning techniques have been developed to adapt LLMs to specific tasks \cite{hu2022lora, dettmers2023qlora, liu2024dora, sun2025dynamic}. Prior work has successfully applied these methods to domains such as legal document drafting \cite{lin2024legal}, medical diagnosis \cite{wu2023medical}, smart contract auditing \cite{ma2024combining}, and tool usage \cite{schick2023toolformer}. However, fine-tuning LLMs for cybersecurity applications remains underexplored.

In this work, we investigate fine-tuning strategies to embed cybersecurity domain knowledge into a foundational LLM. We compare Supervised Fine-Tuning (SFT) \cite{ziegler2019fine}, Low-Rank Adaptation (LoRA) \cite{hu2021lora}, and Quantized Low-Rank Adaptation (QLoRA) \cite{dettmers2024qlora} for their efficacy and efficiency using a publicly available cybersecurity Q\&A dataset. Our results demonstrate that SFT, LoRA, and QLoRA significantly outperform zero-shot inference with the foundational model. Notably, LoRA and QLoRA achieve comparable performance to SFT with substantially lower computational costs. We open-source our implementation to support future research in this area.\footnote{\hyperlink{https://github.com/1000111-1000111/llm-cybersecurity}{https://github.com/1000111-1000111/llm-cybersecurity}.}

\section{Related Work}

\subsection{Large Language Models (LLMs)}
Language generation, as a sequential task, was historically addressed by recurrent architectures like Long Short-Term Memory (LSTM) networks \cite{wen2015semantically, wu2024deep, song2016lstm}. However, the sequential nature of these models hindered large-scale parallelization during training, limiting their capacity. The introduction of the Transformer architecture \cite{vaswani2017attention} marked a paradigm shift by replacing recurrence with self-attention mechanisms, enabling massive parallelization while preserving the ability to model sequential dependencies. Subsequent exploration led to the dominance of the decoder-only architecture for generative tasks \cite{radford2018improving} and the encoder-only architecture for language understanding tasks 

Modern LLMs are typically pre-trained on vast, general-domain corpora using a next-token prediction objective \cite{radford2019language}, which imbues them with broad, generalist capabilities \cite{brown2020language}. Following pre-training, these models often undergo a post-training alignment phase. This includes Supervised Fine-Tuning (SFT) on high-quality instruction datasets to improve their ability to follow commands \cite{ziegler2019fine}, and Reinforcement Learning from Human Feedback (RLHF) to enhance safety and align their behavior with human preferences \cite{ouyang2022training}. The resulting foundation models are capable of performing a wide array of downstream tasks with remarkable zero-shot or few-shot performance \cite{abbasiantaeb2024let, carta2023iterative, kojima2022large}.

Despite their scale, LLMs possess a finite information capacity constrained by their parameter count. This limitation means they cannot achieve optimal performance across all specialized domains, particularly those requiring deep, niche expertise \cite{huang2024compression}. Consequently, fine-tuning remains a critical step to adapt these generalist models and enhance their performance on specific tasks \cite{lin2024legal, wu2023medical, ma2024combining, schick2023toolformer}.

\subsection{Parameter-Efficient Fine-Tuning (PEFT)}
Fully fine-tuning an LLM, which involves updating all of its billions of parameters, is prohibitively expensive in terms of computational resources, memory, and time. This has led to the development of Parameter-Efficient Fine-Tuning (PEFT) methods, which aim to adapt LLMs by updating only a small fraction of their total parameters, thus making the fine-tuning process more accessible and efficient. PEFT techniques can be broadly categorized into several approaches.

One major branch of PEFT involves selective fine-tuning, where only a small, pre-defined subset of the model's original parameters are updated. For instance, a common baseline approach is to freeze the entire model except for the final prediction layer. A more sophisticated example is BitFit, which proposes fine-tuning only the bias terms of the model and leaving all other parameters frozen \cite{zaken2021bitfit}. These methods are efficient as they don't introduce any new parameters and only require storing the updates for a small fraction of the existing ones.

Other strategies focus on model compression. Knowledge distillation involves training a smaller student model to replicate the output distribution of a larger teacher model, effectively transferring the specialized knowledge into a more compact form \cite{hinton2015distilling, sanh2019distilbert}. Pruning techniques aim to reduce model size by identifying and removing redundant or unimportant weights from the network after fine-tuning \cite{lecun1989optimal, han2015learning}.

Among the most successful and widely adopted PEFT strategies are low-rank methods. This approach is motivated by the observation that the weight updates during adaptation have a low intrinsic rank \cite{aghajanyan2020intrinsic}. Instead of modifying the full weight matrix, methods like Low-Rank Adaptation (LoRA) \cite{hu2021lora} freeze the original model weights and inject trainable low-rank matrices into the Transformer layers. By decomposing the weight update into two smaller matrices, LoRA can achieve performance comparable to full fine-tuning while updating only a minuscule fraction (e.g., <0.01\%) of the parameters. This efficiency has made it a cornerstone of modern LLM adaptation. Building on this, subsequent work like QLoRA has combined low-rank adaptation with quantization techniques to further reduce memory usage, enabling the fine-tuning of massive models on a single GPU \cite{dettmers2024qlora}.

\section{Methodology}
In our study, we selected Llama 3 \cite{dubey2024llama} as the foundational LLM. We fine-tuned this model on the CyberMetric-10000 dataset \cite{tihanyi2024cybermetric}, a collection of cybersecurity-focused question-answer pairs. The dataset was partitioned into a training set (80\%) and an evaluation set (20\%). We evaluated three distinct fine-tuning strategies: Supervised Fine-Tuning (SFT), Low-Rank Adaptation (LoRA), and Quantized Low-Rank Adaptation (QLoRA) to assess their effectiveness and computational efficiency.

\subsection{SFT}
Supervised Fine-Tuning (SFT) is a straightforward yet powerful method for adapting a pre-trained LLM to a specific domain. This process involves further training the model on a labeled, high-quality dataset, updating all of its parameters to align its behavior with the target distribution. SFT has been established as a common practice for model alignment, as demonstrated in the development of models like InstructGPT \cite{ouyang2022training}. While highly effective, SFT is computationally intensive, requiring significant memory and processing power to update the entire parameter space of the model, which can consist of billions of parameters.

\subsection{Low-Rank Adaptation (LoRA)}
The high computational cost of SFT has spurred the development of PEFT methods. LoRA is a prominent PEFT technique that significantly reduces the number of trainable parameters \cite{hu2021lora}. LoRA is based on the hypothesis that the change in weights during model adaptation has a low "intrinsic rank." Instead of updating the original weight matrix $W_0$, LoRA freezes $W_0$ and injects a trainable rank decomposition matrix, $\Delta W$, into the layers of the Transformer architecture.

For a pre-trained weight matrix $W_0 \in \mathbb{R}^{d \times k}$, the update is represented by two smaller matrices, $B \in \mathbb{R}^{d \times r}$ and $A \in \mathbb{R}^{r \times k}$, where the rank $r \ll \min(d, k)$. During training, only matrices $A$ and $B$ are updated, and the new weight matrix is computed as $W = W_0 + BA$. This approach drastically reduces memory requirements, as the number of trainable parameters is only the sum of the parameters in $A$ and $B$. For our experiments, we configured LoRA with a rank of $r=8$.

\subsection{Quantized Low-Rank Adaptation (QLoRA)}
QLoRA further enhances the efficiency of LoRA by incorporating quantization, a technique that reduces the memory footprint of a model by representing its weights with lower-precision data types \cite{dettmers2024qlora}. This method makes fine-tuning large models even more accessible and computationally feasible.

Specifically, QLoRA freezes the foundational model's weights in a quantized 4-bit NormalFloat (NF4) format. The LoRA adapter matrices, which are kept in a higher-precision format, are then fine-tuned on top of these quantized base model weights. QLoRA also introduces double quantization, a method that quantizes the quantization constants themselves, yielding additional memory savings without a significant drop in performance. This combination of quantization and low-rank adaptation makes it feasible to fine-tune massive models on a single, consumer-grade GPU, dramatically improving the accessibility of LLM adaptation.

\section{Experiments}

\subsection{Training}
We conducted fine-tuning on the Llama 3 base model using three distinct methods: full SFT, LoRA, and QLoRA. The specific hyperparameters for each training run are detailed in Table \ref{tab:training_params}.

\begin{table}[ht]
\renewcommand{\arraystretch}{1.5}
\centering
\begin{tabular}{|p{4cm}|p{2.5cm}|p{2.5cm}|p{2.5cm}|}
\hline
\textbf{Training Parameter}       & \textbf{SFT} & \textbf{LoRA} & \textbf{QLoRA}    \\ \hline
\textbf{Number of Ranks}           & Full          & 8             & 8               \\ \hline
\textbf{Quantization}              & N/A           & N/A           & 4-bit Quantized \\ \hline
\textbf{Training Batch Size}       & 1             & 4             & 4               \\ \hline
\textbf{Gradient Accumulation}     & 2             & 1             & 1               \\ \hline
\textbf{Learning Rate}             & 7e-6          & 1e-4          & 1e-4            \\ \hline
\textbf{Cutoff Length (Tokens)}    & 1600          & 1600          & 1600            \\ \hline
\textbf{Number of Epochs}          & 3.0           & 3.0           & 3.0             \\ \hline
\textbf{Scheduler}                 & Cosine        & Cosine        & Cosine          \\ \hline
\textbf{Warm-up Ratio}              & 0.1           & 0.1           & 0.1             \\ \hline
\end{tabular}
\caption{Training Parameters for SFT, LoRA, and QLoRA.}
\label{tab:training_params}
\end{table}

For SFT, we updated the full parameter set of the model. Due to the high memory requirements of this approach, we used a small training batch size of 1 with a gradient accumulation of 2, resulting in an effective batch size of 2. A conservative learning rate of $7 \times 10^{-6}$ was used to ensure stable convergence. For the parameter-efficient methods, LoRA and QLoRA, we set the adapter rank to 8. These methods allowed for a larger batch size of 4 without gradient accumulation. A higher learning rate of $1 \times 10^{-4}$ was employed, which is standard for PEFT methods. The QLoRA run further leveraged 4-bit quantization to reduce the memory footprint. All experiments were run for 3 epochs with a cosine learning rate scheduler and a warm-up ratio of 0.1 to stabilize training in the initial phase. The input sequence length was capped at 1600 tokens for all runs.

\begin{figure}[ht]
    \centering
    \includegraphics[width=\textwidth]{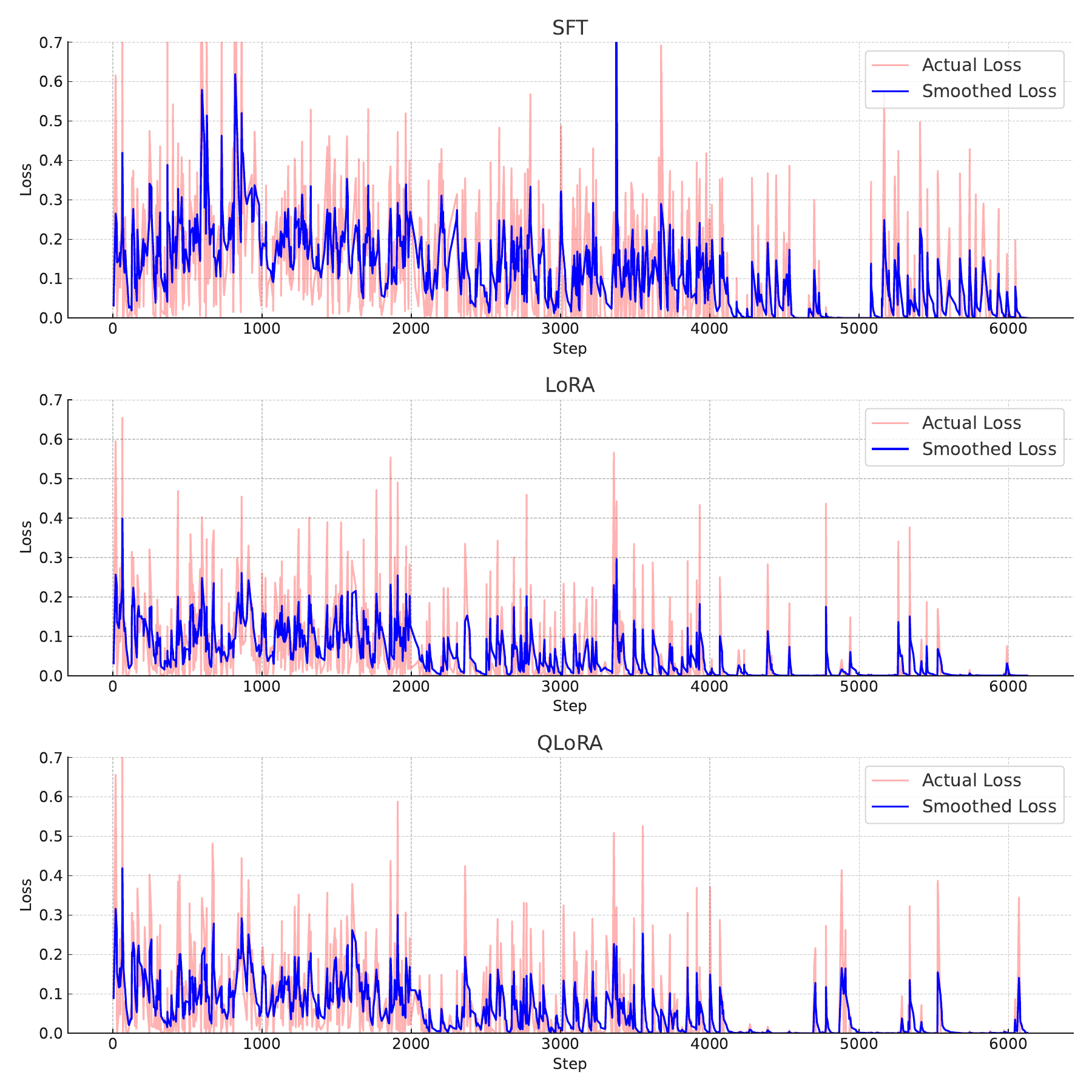}
    \caption{Training losses for SFT, LoRA, and QLoRA. The semi-transparent lines represent actual loss values, while the solid lines represent smoothed values.}
    \label{fig:loss_comparison}
\end{figure}

The training loss curves for all three methods are illustrated in Figure \ref{fig:loss_comparison}. The plot shows that all models converged successfully over the training period. The SFT loss curve exhibits slightly more variance compared to the smoother curves of LoRA and QLoRA, which can be attributed to the larger number of parameters being updated.

\subsection{Evaluation}

To assess the performance of each fine-tuned model, we used a comprehensive set of metrics. For evaluating the correctness of the answers, we calculated Accuracy, Precision, Recall, and the F1-score. These metrics are defined as follows, where TP, FP, and FN represent True Positives, False Positives, and False Negatives, respectively:

\begin{align}
    \text{Precision} &= \frac{\text{TP}}{\text{TP} + \text{FP}} \\
    \text{Recall} &= \frac{\text{TP}}{\text{TP} + \text{FN}} \\
    F_1 &= 2 \cdot \frac{\text{Precision} \cdot \text{Recall}}{\text{Precision} + \text{Recall}}
\end{align}
To measure the quality and similarity of the generated text against the ground-truth answers, we also employed BLEU-4, which measures n-gram precision, and ROUGE (ROUGE-1, ROUGE-2, and ROUGE-L), which measures n-gram recall.

\begin{table}[ht]
\renewcommand{\arraystretch}{1.5}
\centering
\begin{tabular}{|p{4cm}|p{2.5cm}|p{2.5cm}|p{2.5cm}|}
\hline
\textbf{Metric}                    & \textbf{SFT}   & \textbf{LoRA}  & \textbf{QLoRA} \\ \hline
\textbf{Accuracy}                  & 0.7621         & 0.8404         & 0.8429         \\ \hline
\textbf{BLEU-4}                    & 96.7491        & 97.8193        & 97.8528        \\ \hline
\textbf{F1}                        & 0.7620         & 0.8404         & 0.8429         \\ \hline
\textbf{Precision}                 & 0.7622         & 0.8404         & 0.8434         \\ \hline
\textbf{Recall}                    & 0.7619         & 0.8404         & 0.8428         \\ \hline
\textbf{ROUGE-1}                   & 76.2115        & 84.0431        & 84.2878        \\ \hline
\textbf{ROUGE-2}                   & 0.0            & 0.0            & 0.0            \\ \hline
\textbf{ROUGE-L}                   & 76.2115        & 84.0431        & 84.2878        \\ \hline
\textbf{Steps per second}          & 6.478          & 6.504          & 6.503          \\ \hline
\end{tabular}
\caption{Evaluation results for SFT, LoRA, and QLoRA.}
\label{tab:eval_metrics}
\end{table}

Our evaluation, summarized in Table \ref{tab:eval_metrics}, reveals several key findings. The foundational Llama 3 model without any fine-tuning achieved an accuracy of only 0.37, establishing a clear baseline. All three fine-tuning methods dramatically improved upon this. The SFT model achieved an accuracy of 0.7621, while LoRA and QLoRA performed even better, reaching accuracies of 0.8404 and 0.8429, respectively. This trend of LoRA and QLoRA outperforming SFT is consistent across F1, Precision, and Recall scores, suggesting that the parameter-efficient approaches were more effective at specializing to the cybersecurity domain.

Notably, the performance of QLoRA is nearly identical to that of LoRA, indicating that 4-bit quantization successfully reduced computational costs without sacrificing model quality. The training throughput, measured in steps per second, was also comparable across all methods. A peculiar result is the ROUGE-2 score of 0.0 for all models, which implies that no model generated a single overlapping two-word sequence with the reference answers. This is likely due to the highly specific nature of the cybersecurity Q\&A dataset, where answers must be exact.

\begin{figure}[ht]
    \centering
    \includegraphics[width=\textwidth]{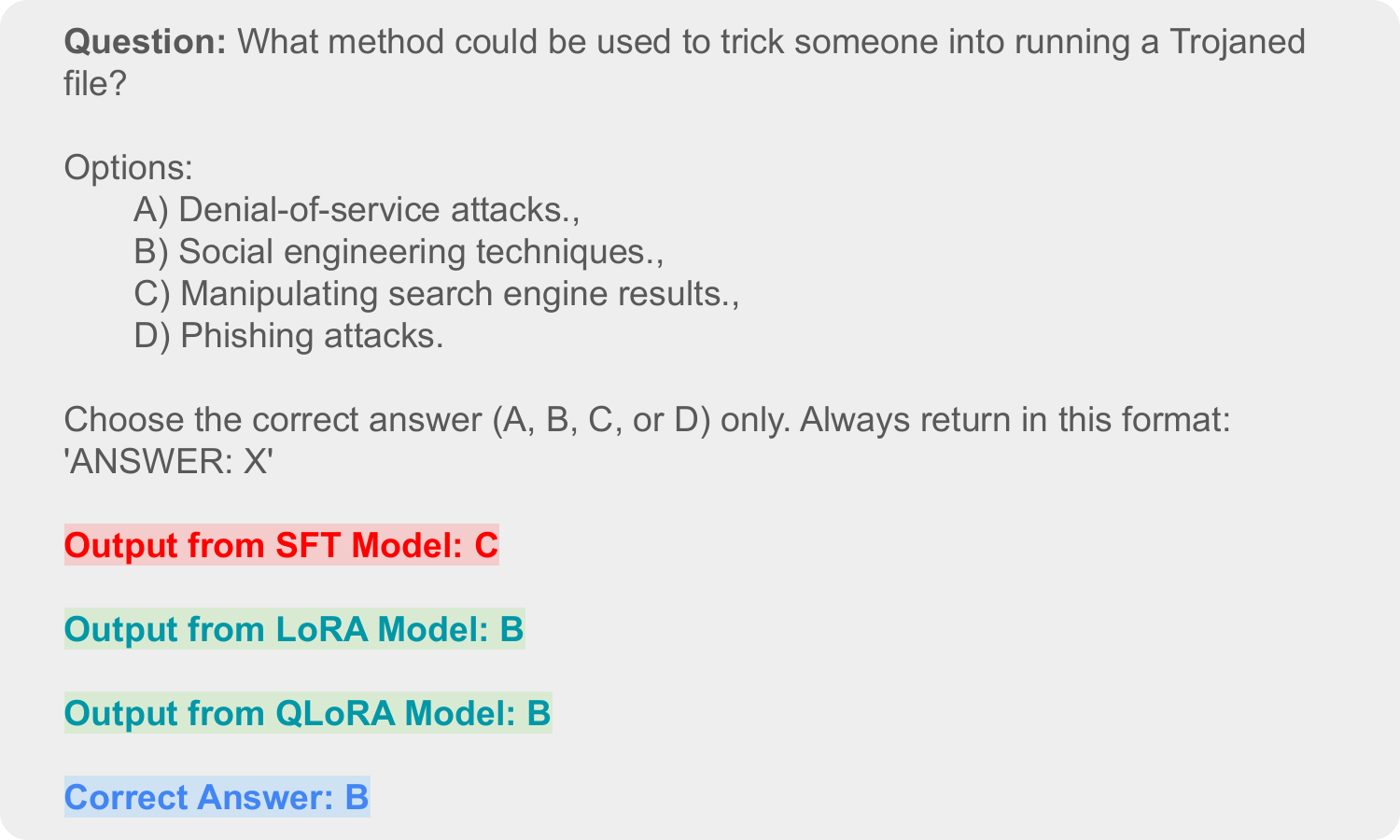}
    \caption{An example question prompt for the fine-tuned models. Both LoRA and QLoRA models generated the correct answer, while the SFT model produced an incorrect one.}
    \label{fig:example_answer}
\end{figure}

\section{Discussion}

\subsection{Conclusion}
In this work, we investigated the efficacy of different fine-tuning strategies for adapting LLMs to the specialized domain of cybersecurity. We compared full SFT with two parameter-efficient methods, LoRA and QLoRA, on a cybersecurity Q\&A task. Our experimental results clearly demonstrate that all fine-tuning methods significantly enhance the model's performance compared to the zero-shot capabilities of the foundational Llama 3 model.

Crucially, our findings highlight that the low-rank methods, LoRA and QLoRA, not only achieve comparable—and in our case, superior—performance to full SFT but do so with substantially greater computational efficiency. The ability of QLoRA to match the performance of LoRA while leveraging 4-bit quantization further underscores the viability of these techniques for democratizing access to LLM customization. By successfully embedding domain-specific knowledge with minimal resource expenditure, LoRA and QLoRA present an efficient and effective pathway for bridging the gap between general-purpose LLMs and the demands of specialized fields like cybersecurity.

\subsection{Future Work}
Building on the promising results of this study, several avenues for future research warrant exploration. First, while our experiments focused on a single foundational model, future work could involve applying these fine-tuning techniques to a broader range of LLM architectures and sizes to assess the generalizability of our findings.

Second, the performance of any fine-tuned model is heavily dependent on the quality and scope of the training data. Expanding the CyberMetric-10000 dataset or creating new, more diverse cybersecurity datasets that encompass a wider variety of tasks—such as threat report analysis, malware classification, or security policy generation—would be a valuable contribution to the field.

Furthermore, the landscape of PEFT is rapidly evolving. Investigating more advanced low-rank methods or other emerging PEFT techniques could yield even greater efficiency and performance gains. Finally, future research should focus on more rigorous, real-world evaluations. This could involve deploying these fine-tuned models in practical cybersecurity environments or developing more challenging benchmarks that test for deeper reasoning and problem-solving capabilities beyond structured Q\&A.

\bibliography{references}

\begin{thebibliography}{10}

\bibitem{vaswani2017attention}
A.~Vaswani, ``Attention is all you need,'' {\em Advances in Neural Information Processing Systems}, 2017.

\bibitem{kaplan2020scaling}
J.~Kaplan, S.~McCandlish, T.~Henighan, T.~B. Brown, B.~Chess, R.~Child, S.~Gray, A.~Radford, J.~Wu, and D.~Amodei, ``Scaling laws for neural language models,'' {\em arXiv preprint arXiv:2001.08361}, 2020.

\bibitem{achiam2023gpt}
J.~Achiam, S.~Adler, S.~Agarwal, L.~Ahmad, I.~Akkaya, F.~L. Aleman, D.~Almeida, J.~Altenschmidt, S.~Altman, S.~Anadkat, {\em et~al.}, ``Gpt-4 technical report,'' {\em arXiv preprint arXiv:2303.08774}, 2023.

\bibitem{abbasiantaeb2024let}
Z.~Abbasiantaeb, Y.~Yuan, E.~Kanoulas, and M.~Aliannejadi, ``Let the llms talk: Simulating human-to-human conversational qa via zero-shot llm-to-llm interactions,'' in {\em Proceedings of the 17th ACM International Conference on Web Search and Data Mining}, pp.~8--17, 2024.

\bibitem{carta2023iterative}
S.~Carta, A.~Giuliani, L.~Piano, A.~S. Podda, L.~Pompianu, and S.~G. Tiddia, ``Iterative zero-shot llm prompting for knowledge graph construction,'' {\em arXiv preprint arXiv:2307.01128}, 2023.

\bibitem{kojima2022large}
T.~Kojima, S.~S. Gu, M.~Reid, Y.~Matsuo, and Y.~Iwasawa, ``Large language models are zero-shot reasoners,'' {\em Advances in neural information processing systems}, vol.~35, pp.~22199--22213, 2022.

\bibitem{huang2024compression}
Y.~Huang, J.~Zhang, Z.~Shan, and J.~He, ``Compression represents intelligence linearly,'' {\em arXiv preprint arXiv:2404.09937}, 2024.

\bibitem{hu2022lora}
E.~J. Hu, Y.~Shen, P.~Wallis, Z.~Allen-Zhu, Y.~Li, S.~Wang, L.~Wang, W.~Chen, {\em et~al.}, ``Lora: Low-rank adaptation of large language models.,'' {\em ICLR}, vol.~1, no.~2, p.~3, 2022.

\bibitem{dettmers2023qlora}
T.~Dettmers, A.~Pagnoni, A.~Holtzman, and L.~Zettlemoyer, ``Qlora: Efficient finetuning of quantized llms,'' {\em Advances in neural information processing systems}, vol.~36, pp.~10088--10115, 2023.

\bibitem{liu2024dora}
S.-Y. Liu, C.-Y. Wang, H.~Yin, P.~Molchanov, Y.-C.~F. Wang, K.-T. Cheng, and M.-H. Chen, ``Dora: Weight-decomposed low-rank adaptation,'' in {\em Forty-first International Conference on Machine Learning}, 2024.

\bibitem{sun2025dynamic}
X.~Sun, S.~Yang, Y.~Chen, F.~Fan, Y.~Liang, and D.~Rakita, ``Dynamic rank adjustment in diffusion policies for efficient and flexible training,'' {\em arXiv preprint arXiv:2502.03822}, 2025.

\bibitem{lin2024legal}
C.-H. Lin and P.-J. Cheng, ``Legal documents drafting with fine-tuned pre-trained large language model,'' {\em arXiv preprint arXiv:2406.04202}, 2024.

\bibitem{wu2023medical}
C.~Wu, Z.~Lin, W.~Fang, and Y.~Huang, ``A medical diagnostic assistant based on llm,'' in {\em China Health Information Processing Conference}, pp.~135--147, Springer, 2023.

\bibitem{ma2024combining}
W.~Ma, D.~Wu, Y.~Sun, T.~Wang, S.~Liu, J.~Zhang, Y.~Xue, and Y.~Liu, ``Combining fine-tuning and llm-based agents for intuitive smart contract auditing with justifications,'' {\em arXiv preprint arXiv:2403.16073}, 2024.

\bibitem{schick2023toolformer}
T.~Schick, J.~Dwivedi-Yu, R.~Dess{\`\i}, R.~Raileanu, M.~Lomeli, E.~Hambro, L.~Zettlemoyer, N.~Cancedda, and T.~Scialom, ``Toolformer: Language models can teach themselves to use tools,'' {\em Advances in Neural Information Processing Systems}, vol.~36, pp.~68539--68551, 2023.

\bibitem{ziegler2019fine}
D.~M. Ziegler, N.~Stiennon, J.~Wu, T.~B. Brown, A.~Radford, D.~Amodei, P.~Christiano, and G.~Irving, ``Fine-tuning language models from human preferences,'' {\em arXiv preprint arXiv:1909.08593}, 2019.

\bibitem{hu2021lora}
E.~J. Hu, Y.~Shen, P.~Wallis, Z.~Allen-Zhu, Y.~Li, S.~Wang, L.~Wang, and W.~Chen, ``Lora: Low-rank adaptation of large language models,'' {\em arXiv preprint arXiv:2106.09685}, 2021.

\bibitem{dettmers2024qlora}
T.~Dettmers, A.~Pagnoni, A.~Holtzman, and L.~Zettlemoyer, ``Qlora: Efficient finetuning of quantized llms,'' {\em Advances in Neural Information Processing Systems}, vol.~36, 2024.

\bibitem{wen2015semantically}
T.-H. Wen, M.~Gasic, N.~Mrk{\v{s}}i{\'c}, P.-H. Su, D.~Vandyke, and S.~Young, ``Semantically conditioned lstm-based natural language generation for spoken dialogue systems,'' in {\em Proceedings of the 2015 Conference on Empirical Methods in Natural Language Processing}, pp.~1711--1721, 2015.

\bibitem{wu2024deep}
Y.~Wu, X.~Sun, I.~Spasojevic, and V.~Kumar, ``Deep learning for optimization of trajectories for quadrotors,'' {\em IEEE Robotics and Automation Letters}, vol.~9, no.~3, pp.~2479--2486, 2024.

\bibitem{song2016lstm}
J.~Song, S.~Tang, J.~Xiao, F.~Wu, and Z.~M. Zhang, ``Lstm-in-lstm for generating long descriptions of images,'' {\em Computational Visual Media}, vol.~2, no.~4, pp.~379--388, 2016.

\bibitem{radford2018improving}
A.~Radford, K.~Narasimhan, T.~Salimans, I.~Sutskever, {\em et~al.}, ``Improving language understanding by generative pre-training,'' 2018.

\bibitem{radford2019language}
A.~Radford, J.~Wu, R.~Child, D.~Luan, D.~Amodei, I.~Sutskever, {\em et~al.}, ``Language models are unsupervised multitask learners,'' {\em OpenAI blog}, vol.~1, no.~8, p.~9, 2019.

\bibitem{brown2020language}
T.~Brown, B.~Mann, N.~Ryder, M.~Subbiah, J.~D. Kaplan, P.~Dhariwal, A.~Neelakantan, P.~Shyam, G.~Sastry, A.~Askell, {\em et~al.}, ``Language models are few-shot learners,'' {\em Advances in neural information processing systems}, vol.~33, pp.~1877--1901, 2020.

\bibitem{ouyang2022training}
L.~Ouyang, J.~Wu, X.~Jiang, D.~Almeida, C.~Wainwright, P.~Mishkin, C.~Zhang, S.~Agarwal, K.~Slama, A.~Ray, {\em et~al.}, ``Training language models to follow instructions with human feedback,'' {\em Advances in neural information processing systems}, vol.~35, pp.~27730--27744, 2022.

\bibitem{zaken2021bitfit}
E.~B. Zaken, S.~Ravfogel, and Y.~Goldberg, ``Bitfit: Simple parameter-efficient fine-tuning for transformer-based masked language-models,'' {\em arXiv preprint arXiv:2106.10199}, 2021.

\bibitem{hinton2015distilling}
G.~Hinton, O.~Vinyals, and J.~Dean, ``Distilling the knowledge in a neural network,'' {\em arXiv preprint arXiv:1503.02531}, 2015.

\bibitem{sanh2019distilbert}
V.~Sanh, L.~Debut, J.~Chaumond, and T.~Wolf, ``Distilbert, a distilled version of bert: smaller, faster, cheaper and lighter,'' {\em arXiv preprint arXiv:1910.01108}, 2019.

\bibitem{lecun1989optimal}
Y.~LeCun, J.~Denker, and S.~Solla, ``Optimal brain damage,'' {\em Advances in neural information processing systems}, vol.~2, 1989.

\bibitem{han2015learning}
S.~Han, J.~Pool, J.~Tran, and W.~Dally, ``Learning both weights and connections for efficient neural network,'' {\em Advances in neural information processing systems}, vol.~28, 2015.

\bibitem{aghajanyan2020intrinsic}
A.~Aghajanyan, L.~Zettlemoyer, and S.~Gupta, ``Intrinsic dimensionality explains the effectiveness of language model fine-tuning,'' {\em arXiv preprint arXiv:2012.13255}, 2020.

\bibitem{dubey2024llama}
A.~Dubey, A.~Jauhri, A.~Pandey, A.~Kadian, A.~Al-Dahle, A.~Letman, A.~Mathur, A.~Schelten, A.~Yang, A.~Fan, {\em et~al.}, ``The llama 3 herd of models,'' {\em arXiv preprint arXiv:2407.21783}, 2024.

\bibitem{tihanyi2024cybermetric}
N.~Tihanyi, M.~A. Ferrag, R.~Jain, T.~Bisztray, and M.~Debbah, ``Cybermetric: A benchmark dataset based on retrieval-augmented generation for evaluating llms in cybersecurity knowledge,'' in {\em 2024 IEEE International Conference on Cyber Security and Resilience (CSR)}, pp.~296--302, IEEE, 2024.

\end{thebibliography}

\end{document}